\documentclass[11pt,a4paper]{article}
\usepackage[hyperref]{acl2020}
\usepackage{times}
\usepackage{latexsym}
\usepackage{csquotes}
\usepackage{amsmath}
\usepackage{float}
\usepackage[colorinlistoftodos,prependcaption,textsize=tiny,disable]{todonotes}

\usepackage{microtype}

\aclfinalcopy % Uncomment this line for the final submission
\title{Crypto Pump and Dump Detection via Deep Learning Techniques}

\author{Viswanath Chadalapaka \\
  \texttt{vchad@usc.edu} \\\And
  Kyle Chang \\
  \texttt{krchang@usc.edu} \\\And
  Gireesh Mahajan \\
  \texttt{gmahajan@usc.edu} \\\And
  Anuj Vasil \\
  \texttt{avasil@usc.edu} \\}

\date{May 8, 2022}

\begin{document}
\maketitle
\begin{abstract}
\todo{viswanath} Despite the fact that cryptocurrencies themselves have experienced an astonishing rate of adoption over the last decade, cryptocurrency fraud detection is a heavily under-researched problem area. Of all fraudulent activity regarding cryptocurrencies, pump and dump schemes are some of the most common. Though some studies have been done on these kinds of scams in the stock market, the lack of labelled stock data and the volatility unique to the cryptocurrency space constrains the applicability of studies on the stock market toward this problem domain. Furthermore, the only work done in this space thus far has been either statistical in nature, or has been concerned with classical machine learning models such as random forest trees. We propose the novel application of two existing neural network architectures to this problem domain and show that deep learning solutions can significantly outperform all other existing pump and dump detection methods for cryptocurrencies.
\end{abstract}

\section{Introduction}
\todo{viswanath} In 2021 alone, cryptocurrencies (or \enquote{crypto}) experienced over \$14 trillion worth of trading volume, representing nearly a 700\% increase over the previous year. In that same year, the Binance cryptocurrency exchange (or just \enquote{exchange}) was responsible for over two-thirds of that volume \cite{cryptostats_khatri_2021}. As crypto and the use of these exchanges enters the mainstream, the conversation surrounding their regulatory hurdles has intensified \cite{regulation_jrfm12030126}. Of those hurdles, the detection of fraudulent activities at scale is one of the most pressing, given the rapid growth of the space. Despite these concerns, the amount of regulation in crypto pales in comparison to more mainstream flows of money, such as stocks. This allows fraud to plague the crypto space at a level unheard of in the stock market.

Among all types of fraud in the crypto space, pump and dump (P\&D) schemes are some of the most popular, and some of the easiest, to execute on \cite{twomey2020fraud} -- generally resulting from the concerted effort of just a single online P\&D planning group \cite{hamrick}. Despite that, no studies have been done to-date on the application of deep learning to P\&D detection in crypto. Though some studies involving the application of deep learning have been done in the context of traditional securities such as stocks \cite{stock_pandd_8628777}, these studies not only lacked the amount of data now freely available for crypto (via the blockchain), but also have no guarantee of applicability to the much more volatile world of crypto.

This paper presents two novel applications of existing deep learning methods to detect P\&D schemes in crypto -- specifically, for small, volatile cryptocurrencies also known as altcoins. Our work focuses on taking advantage of the bulk of freely available data by using deep learning to drive performance gains, since previous works in this space have thus far only applied either classical machine learning techniques such as random forest trees \cite{laMorgia9209660}, or more basic techniques such as statistic analyses \cite{kamps} to the problem. All of our work is reproducible, and a link to our code can be found in the footnote below.
\footnote{\url{https://github.com/Derposoft/crypto_pump_and_dump_with_deep_learning}}

\begin{figure*}[ht]
\title{Example of a P\&D Scheme on REAP Token}
\begin{center}
\includegraphics[scale=2.0,trim=100 0 0 1,clip]{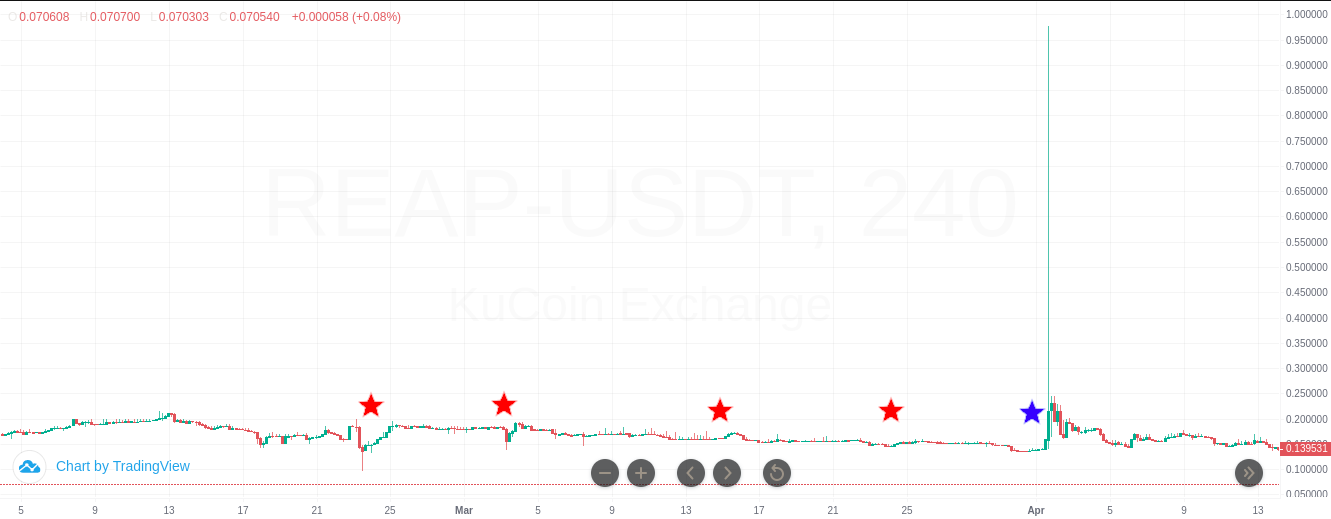}
\end{center}
\caption{An example of an organized P\&D scheme on the REAP token. Red stars indicate visual signs of the accumulation phase in the form of subtle buying pressure, while the blue star indicates the pump and dump phases of the scheme. This data was manually obtained directly from the KuCoin exchange (https://www.kucoin.com/trade/REAP-USDT), although the same information can be found on other exchanges as well. Publicly-available videos on YouTube can confirm that this fluctuation in the REAP token's price was a result of a planned P\&D operation \cite{coffeezilla_2021}. }
\label{pnd_graph}
\end{figure*}

\section{Related Work}
\todo{anuj} 
\todo{viswanath}
\subsection{Application of Classical Machine Learning and Statistical Models to Crypto P\&D Detection}
To-date, only classical machine learning models such as random forest trees have been applied in the crypto space to detect P\&D schemes \cite{laMorgia9209660}. Statistical models have also been applied in this domain \cite{kamps}. These models often aggregate trade data at a high level and use these coalesced features to predict when a P\&D scheme will occur. As these are the only works that have been done in this niche, we hope that our own work in this paper will help build a better foundation in this domain going forward.

\subsection{Application of Deep Learning to General Anomaly Detection}

Anomaly detection in a general setting unrelated to crypto is a well-researched field. More specifically, studies involving the application of deep learning to time-series anomaly detection problems have been done involving multiple network architectures: LSTMs \cite{malhotra2016lstm}, convolutional networks \cite{kwon2018empiricalconv}, and various combinations of the two
\cite{clstmKIM201866}; and more recently, multiple variations of attention-based methods such as RNN attention \cite{transformerbrown2018recurrent} or the anomaly transformer \cite{anomalytransformer} have also been explored. As of writing this paper, deep learning architectures are at the forefront of time-series anomaly detection due to their strength in making predictions using spatiotemporal relationships, which are key to strong anomaly detection models \cite{clstmKIM201866}. In this work, we implement, modify, and tune some of these latest architectures in an attempt to adapt them to the crypto domain.

\section{Background}
\todo{anuj}
While P\&D is heavily fined by the SEC in traditional financial markets (e.g. stocks), crypto remains to be regulated at a federal level. This opens the opportunity for various P\&D groups to openly plan P\&D events over online messaging platforms such as Telegram or Discord \cite{laMorgia9209660}. Typically, a P\&D scheme consists of three phases: the accumulation phase, the promotion (or \enquote{pump}) phase, and the distribution (or \enquote{dump}) phase.

During the accumulation phase, the group organizing the scheme slowly accumulates a significant position in the asset of interest. Once ready, the promotion phase begins: excitement is drummed up via social media promotion for the asset, and bullish sentiment is falsified through fraudulent reports. This causes retail investors to rally behind and invest in the asset, thereby driving the price up. Finally, during the distribution phase, the perpetrator of the scheme liquidates their position in the asset over a very short period of time. Since the position being liquidated generally represents a significant portion of the asset itself, this inevitably causes a crash in the price of the original asset -- leaving most of the retail investors who bought during the promotion phase at a significant loss on their positions over an extremely short time.

The graph in Fig.~\ref{pnd_graph} is an example of a real P\&D scheme on the REAP token which occurred near the start of April 2021. %The red stars near the start of the graph show subtle signs of buying pressure, and are a visual signal of an accumulation phase on the coin. The blue star near the start of April signals both the pump and the dump phase of the coin -- i
In this case, the entirety of the pump and dump phases lasted only a few minutes, and the accumulating party dumped all of their coins as soon as the value of the token shot up. Investors who bought during the peak of the pump would have been down nearly 90\% in the span of just a few minutes.

\iffalse
\begin{figure*}[ht]
\title{Example of a P\&D Scheme on REAP Token}
\begin{center}
\includegraphics[scale=1.0,trim=100 0 0 40,clip]{pumpanddump_annotated.png}
\end{center}
\label{pnd_graph}
\caption{test}
\end{figure*}
\fi

Successful P\&D detection algorithms have the potential to alert investors of P\&D schemes before they occur, and could even enable regulatory bodies to police this fraudulent activity in more cost and time-effective manners. 
Ultimately, this could help to establish crypto as a much more legitimate trading option, as well as save investors and exchanges millions from potential scams and fraudulent activity.

\section{Method}
\todo{viswanath} We propose the novel application of two architectures that have scored well on standard anomaly detection datasets to the now-burgeoning financial data also available in the cryptocurrency space: the C-LSTM model \cite{clstmKIM201866}, and the Anomaly Transformer model \cite{anomalytransformer}.

Since P\&D schemes often have multiple phases that generally occur over vastly different lengths of time (e.g. the accumulation phase may last for up to a month, whereas the pump or dump phases can last for as little as a minute), both of the models that we have chosen have the ability to capture both longer-term anomalies -- otherwise known as \enquote{trend} anomalies -- and much shorter-term anomalies -- otherwise known as \enquote{point} anomalies. This capability is important for the detection of P\&D schemes, since models that are only capable of one or the other could potentially be fooled by the volatility inherent in crypto markets.

\subsection{Models}
\subsubsection{C-LSTM}
\todo{viswanath} Our first model is the C-LSTM model, originally introduced by \cite{clstmKIM201866} for learning anomaly detection by treating data as spatiotemporal in nature. The model consists of a series of convolutional/ReLU/pooling layers to encode the input sequence, followed by a set of LSTM layers, with decoding done via a set of feedforward layers. In this model, convolutional layers help to capture spatial information within the dataset, thereby helping to identify point anomalies; on the other hand, the LSTM layers help to capture temporal information, and help to identify trend anomalies. This simple model has shown success in the detection of varying types of web traffic anomalies \cite{clstmKIM201866}, as well as anomalies across a number of stocks in the Chinese stock market \cite{yang9382507}. The following image is a visual representation of the C-LSTM model.
\begin{center}
\includegraphics[scale=0.155,trim= 0 420 0 335,clip]{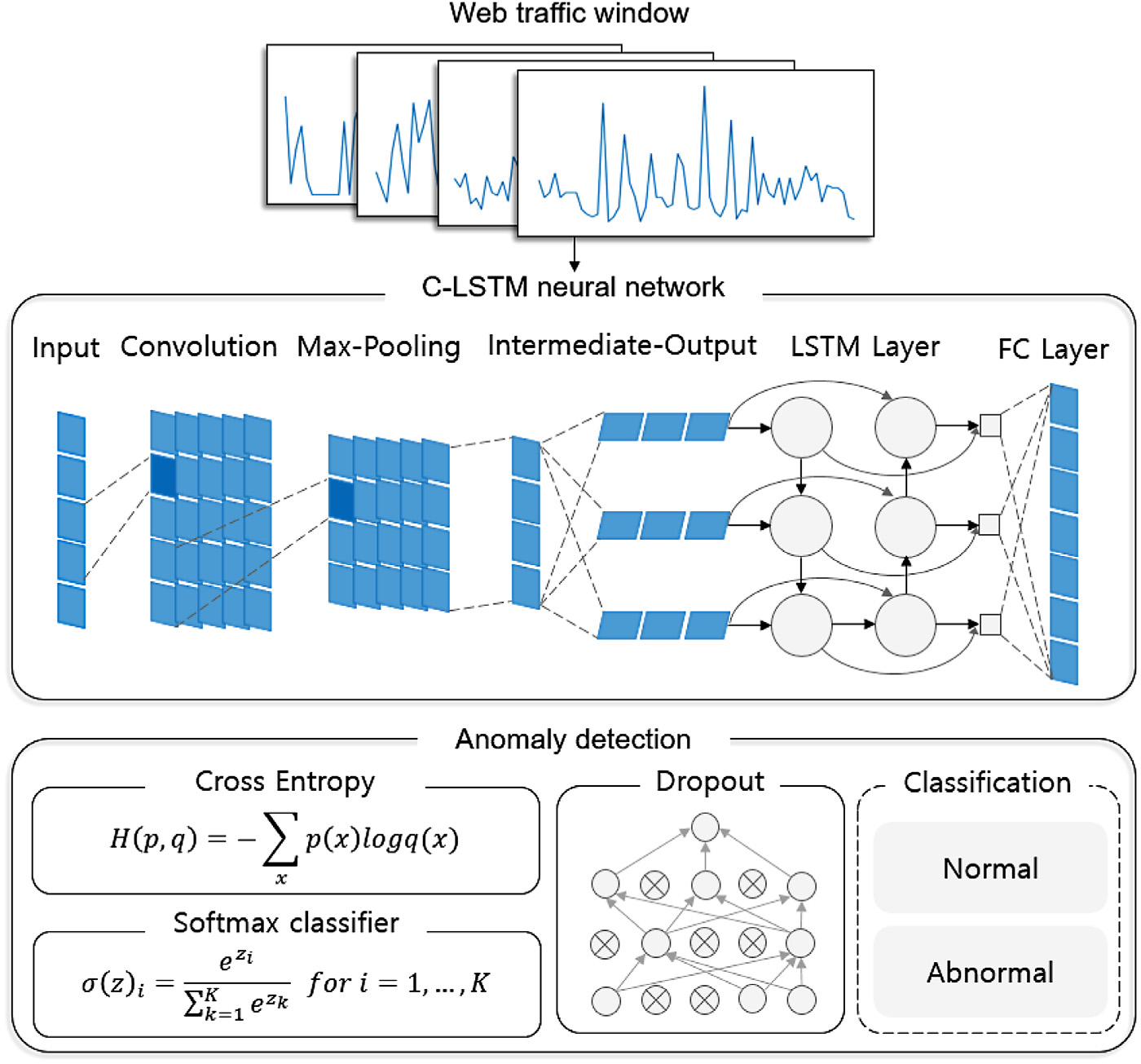}
\end{center}
\paragraph{Our Implementation}
Our own model uses the following architecture: 1 set of convolutional/ReLU/pooling layers with a convolution kernel size of 3 with a stride of 1, and a pooling kernel size of 2 with a stride of 1; 1 LSTM layer with an embedding dimension of 350; 1 feedforward layer which directly projects the last hidden state of the LSTM to a dimension of 1; and a sigmoid layer which constrains the output of our classifier between 0 and 1.

This setup resulted in a C-LSTM model with 997,851 learnable parameters which was trained for 200 epochs.

\subsubsection{Anomaly Transformer}
\todo{Gireesh}
The second model we explored was the Anomaly Transformer introduced by \cite{anomalytransformer}. Unlike a standard transformer, the Anomaly Transformer uses a custom anomaly attention module to improve its performance in anomaly detection scenarios. To that end, the model introduces two novelties: an anomaly attention module, and a minimax optimization strategy. Additionally, on top of this model, we introduced our own, original novelty -- an altered optimization strategy and loss function -- in order to adapt the model to a supervised setting, since this model was first developed and intended for unsupervised settings. The Anomaly Transformer model was chosen because it is the current state-of-the-art anomaly detection model for general time-series anomaly detection \cite{anomalytransformer}, consistently achieving strong results across a variety of standard datasets, including server sensor data \cite{10.1145/3292500.3330672}, rover data from NASA \cite{nasadataset}, and the NeurIPS 2021 time-series benchmark \cite{neuripstimeseriesbenchmark}.

The first of the two novelties introduced by the Anomaly Transformer which makes it particularly good at detecting anomalies is the replacement of the standard self-attention computation with two, internally-computed association values that influence the attention block: series association and prior association. These two associations make up the \enquote{anomaly attention module} and are as described below:
\begin{enumerate}
    \item \textbf{Series association} at each layer $l$, denoted $S^l$, is a simple self-attention computation on the data before multiplying by the value matrix $V$ which is standard to a normal attention mechanism. Series association focuses on longer-term trends, and helps in the detection of trend anomalies.
\[S^l = Softmax(\frac{QK^T}{\sqrt{d_{model}}})\]
    \item \textbf{Prior association} at each layer $l$, denoted $P^l$, computes a Gaussian kernel for each point with respect to the input sequence. A rescale operation is done to convert the result to discrete distributions for each layer $l$. Prior association focuses more on local points, and helps the model in the detection of point anomalies.
%\[P^l = Rescale\left(\left[\frac{1}{\sqrt{2\pi}\sigma_i}\exp\Big(-\frac{|j-i|^2}{2\sigma^2_i}\Big)\right]_{i,j \in {1,...,N}}\right)\]

\[P^l = Rescale\left(\left[\mathcal{N}(j | \mu=i, \sigma)\right]_{i,j \in {1,...,N}}\right)\]
\end{enumerate}

The second of the two novelties introduced by this model is the minimax optimization process which takes place after computing the values $S^l$ and $P^l$. This process consists of two phases: minimize and maximize. During the minimize phase, the series association is moved towards the prior association using the symmetric KL divergence formula, denoted $SKL$. During the maximize phase, the series association is moved towards the original input sequence by taking the absolute difference of the two. The series association is also enlarged using the prior association (once again through the symmetric KL divergence formula). Therefore, during the minimize phase, the prior association approximates the series association. Then, during the maximize phase, the series association pays more attention to non adjacent points since it is enlarged by the prior association. As a result, a sequence generated by prior and series associations will lower the attention value at any anomaly. The total amount of difference between the two associations captured during this process is encapsulated by the Association Discrepancy function, denoted as $AD$. The data is then reconstructed after the minimax optimization via multiplying the series association $S^l$ with the value matrix standard to an attention module, $V$. The minimax strategy is visualized below.
\includegraphics[scale=0.7]{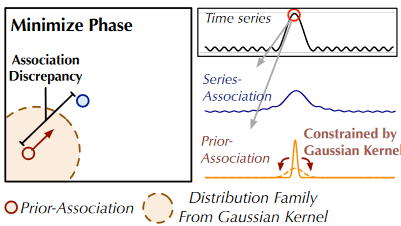}
\includegraphics[scale=0.7]{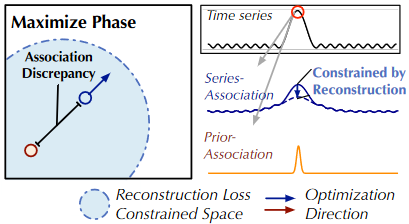}
Combining these novelties means that the reconstructed series, denoted $\hat{X}$, will greatly differ from the original series if an anomaly is present, making the discrepancy greater and giving the model an easier time to identify anomalies. After integrating in the reconstruction loss, the losses derived from each of these two phases are summarized through the following equations, where $\lambda > 0$ and $L_{tot}$ represents the total loss:

%\[Minimize Phase: L_{total}(\hat{X}, P, S_{detach}, -\lambda; X)\]
%\[Maximize Phase: L_{total}(\hat{X}, P, S_{detach}, \lambda; X)\]
%\[L_{total}(\hat{X},P,S,\lambda;X) = \|X - \hax{X}\|^2_F - \lambda \times \| AssDis(P,S;X)\|\]
%\[AssDis(P,S;X) = \left[\frac{1}{L} \sum^L_{l=1} \left( KL(P^l_{i,:} \| S^l_{i,:}) + KL(S^l_{i,:} \| P^l_{i,:}) \right)\right] \]

\[Min Phase: L_{tot}(\hat{X}, P, S_{detach}, -\lambda; X)\]
\[Max Phase: L_{tot}(\hat{X}, P_{detach}, S, \lambda; X)\]
\begin{equation*}
\begin{aligned}
L_{tot}(\hat{X},P,S,\lambda;X) = {} &\|X - \hat{X}\|^2_F \\
&- \lambda \| AD(P,S;X)\|
\end{aligned}
\end{equation*}
\[SKL(A, B) = KL(A \| B) + KL(B \| A)\]
\begin{equation*}
\begin{aligned}
AD(P,S;X) = \frac{1}{L} \sum^L_{l=1} SKL(P^l_{i,:}, S^l_{i,:})
\end{aligned}
\end{equation*}
\[\hat{X} = S^lV\]

\paragraph{Our Implementation}

Since our own problem was supervised, we modified the optimization strategy and loss function used, since the Anomaly Transformer was originally built for unsupervised data. We propose and implement the following changes in order to adapt this model to a supervised anomaly detection setting:
\begin{enumerate}
    \item Keep the maximize phase the same, but replace the minimize phase: instead, we move the series association towards the ground truth.
    \item Replace the first term of the loss with the popular MSE loss between the outputted labels and the ground truth labels.
\end{enumerate}
Our loss function for our maximize phase therefore becomes the following:
\begin{equation*}
\begin{aligned}
L_{tot}(\hat{Y},P,S,\lambda;Y) = {} &MSE(\hat{Y},Y) \\
&- \lambda \| AD(P_{detach},S;\hat{Y})\|
\end{aligned}
\end{equation*}

The optimal model we used had a sequence length of 15, 4 layers, and a lambda of 0.0001.

This setup resulted in an Anomaly Transformer model with 155,665 learnable parameters which was trained for 50 epochs.
\subsection{Metrics}
\todo{viswanath} During our experiments, we collect the precision, recall, and F1 scores of our models on the validation set at the end of each epoch. Since cryptocurrency P\&D detection is an anomaly detection problem, we have chosen not to collect the accuracy of our model, because accuracy measurements do not reflect the strength of a model in anomaly detection scenarios. Specifically, since anomaly detection scenarios have a large class imbalance heavily favoring negative labels, the accuracy of any model is almost always over 99.9\% even if it just outputs 0 (the negative class). Therefore, we optimize our models for the best possible F1 score for the reason that F1 scores generally reflect the strength of an anomaly detection model accurately. This is the standard used by most other anomaly detection models and helps us compare our results to previous work more effectively.

The precision is defined as the percentage of predicted anomalies which were correctly classified, and the recall is defined as the percentage of actual anomalies that were correctly classified as anomalies. The F1 score is the harmonic mean of these two values. Given that both the precision and recall are important in the context of anomaly detection, it stands to reason that the F1 score, which is a mix of the two values, is a good metric with which to compare two different models at the end of the day. The following equations are used to calculate the precision, recall, and F1 scores respectively:

$$Precision = \frac{N_{TP}}{N_{TP}+N_{FP}}$$
$$Recall = \frac{N_{TP}}{N_{TP}+N_{FN}}$$
$$F1 = 2*\frac{Precision * Recall}{Precision+Recall}$$

where the terms $N_{TP}$, $N_{FP}$, and $N_{FN}$ refer to the number of True Positives, False Positives, and False Negatives respectively.

After training each model, we choose the metrics corresponding to the epoch in which the highest validation F1 score was obtained instead of the metrics from the final epoch. This ensures that our metrics reflect a model that hasn't overfit the training data.

\section{Experiments}
\subsection{Dataset}
\todo{viswanath} The dataset used in this paper consists of manually-labelled, raw transaction data from the Binance cryptocurrency exchange, first introduced by \cite{laMorgia9209660}. The transactions of various cryptocurrencies that experienced known occurrences of P\&D were collected.

To generate the dataset, the authors first joined several cryptocurrency P\&D Telegram groups that were well-known for planning and executing on P\&D schemes. Then, over a period of two years, the researchers collected the timestamps for official "pump signals" which were announced in each of these groups by group administrators. Using these timestamps and the Binance API, the authors were able to collect every transaction for the pumped cryptocurrency for up to 1 week preceding and succeeding the pump, depending on what was available to access. In this fashion, 343 P\&D occurrences' worth of data was collected.

After collecting this raw data from the Binance API, the authors further preprocessed the data by aggregating transactions into 5-second, 15-second, and 25-second \enquote{chunks}, thereby forming three different aggregated datasets. Each of these aggregated datasets, in turn, contains the following 15 features:
\begin{enumerate}
    \item \textbf{Date, HourSin, HourCos, MinuteSin, MinuteCos:} The date and hour- and minute-based positional encoding of a given chunk.
    \item \textbf{PumpIndex, Symbol:} The 0-based index of the pump, numbering it out of the 343 available pumps, and the ticker symbol of the coin on which the pump took place.
    \item \textbf{StdRushOrder, AvgRushOrder:} The moving standard deviation and average percent change of the number of rush orders.
    \item \textbf{StdTrades:} The moving standard deviation of the number of trades, both buy and sell.
    \item \textbf{StdVolume, AvgVolume:} The moving standard deviation and average percent change in the order volume.
    \item \textbf{StdPrice, AvgPrice, AvgPriceMax:} The moving standard deviation, average percent change, and average maximum percent change in the price of the asset.
\end{enumerate}

%Date, HourSin, HourCos, MinuteSin, MinuteCos, PumpIndex, Symbol, StdRushOrder, AvgRushOrder, StdTrades, StdVolume, AvgVolume, StdPrice, AvgPrice, AvgPriceMax

\subsection{Implementation Details}
\todo{Kyle} %talk about how we read the data (segmentation, sliding window, differently-chunked datasets), how many epochs we trained our models for, what hardware we trained our models on, and other implementation-y stuff
Given an input data sequence $X=X_1, ..., X_N$, our first step splits the data into a separate train and validation set using an 80:20 ratio without shuffling. Following this split, train data undergoes the following preprocessing steps: First, data is broken into $M$ contiguous subsequences $X=Y_1, .., Y_M$, where $Y_i$ corresponds to the $i$th pump in the dataset as determined by the PumpIndex feature. Separating the data out by pump at this stage ensures that during training time, models aren't fed information from 2 different pumps at the same time, which may hinder their training process. Of these pumps, all pump sequences $Y_i$ with fewer than 100 chunks are discarded, since pumps that only have a few chunks do not make for exemplary training samples. This leaves us with a training subset of $m < M$ pump sequences.

Then, we further prepare the data for each of these pumps by splitting them into segments of size $s$ via taking a sliding window over the chunks of each pump $Y_i$ and adding reflection padding to the start of size $s-1$ so that the resulting count of windows is equal to the original count of chunks. Together, the chunks in each window are considered to be one \enquote{segment.} In this scheme, segments are inputs to our models. The models then predict the probability of a pump occurring during the \textit{last} chunk of the segment. At this point, the segments from all pumps can once again be safely shuffled and batched together without fear of information from one pump \enquote{leaking} into the training data for another pump.

We choose to segment our data for several reasons. First, segmenting and predicting in this way avoids the possibility for models to make predictions based on future values -- in other words, in our setup, models can only predict whether or not a pump will occur based on currently-known information. This ensures that models trained under this scheme have the capability to be deployed in a real-world, realtime anomaly detection scenarios on real exchanges. Additionally, the C-LSTM model incorporates an LSTM layer which faces performance limitations for longer sequence lengths \cite{attention_viswani}. The data for each pump can also span several days, making it difficult to train models directly on a whole pump sequence due to memory constraints. Having a fixed segment length as opposed to the variable lengths of pump data also simplifies data loading.

Finally, segmenting our data into chunks makes it straightforward to perform undersampling on our data, which has been shown to boost model performance in anomaly detection scenarios by addressing the class imbalance -- even in cases where the reduction does not bring the class ratio all the way down to 50:50 \cite{8424689}. Undersampling also naturally speeds up training, since it reduces the number of training samples. We therefore apply undersampling to our train data by choosing to keep only a random subset of proportion $u$ of all segments that do not contain any anomaly labels. At the same time, we unconditionally include all segments that have an anomaly label anywhere.

Validation data undergoes a similar segmenting process, with the following exceptions: small pumps are not thrown out, and no undersampling is performed. This avoids unfairly skewing metrics in favor of our models.

Via hyperparameter tuning, we found that a segment length value of $s=15$ resulted in the best validation performance as measured by F1 score. An undersampling proportion of $u=0.05$ was used for all experiments except for the C-LSTM on the 15-second chunked dataset, in which $u=0.1$ was used. For the C-LSTM model, batch sizes of 1200, 600, and 600 were used on the 5-second, 15-second, and 25-second chunked datasets respectively. For the Anomaly Transformer model, a batch size of 32 was used on all datasets. Likewise, the C-LSTM model uses precision-recall thresholds of 0.5, 0.4, and 0.65 on each the three datasets respectively, while the Anomaly Transformer model uses a threshold of 0.48 on all datasets.

In order to keep results consistent across all models, baseline results were also recomputed following the same train-validation split as our own models.

%Allowing more majority class data to remain also preserves more information compared to more aggressive undersampling.

\subsection{Baselines}
\todo{viswanath} %explain the kamps+la morgia baselines we used for our project
The baseline results that we use to compare our models against are those of the random forest model employed by \cite{laMorgia9209660}. For more context, the results of the statistical model introduced by \cite{kamps} are also included in our results table, since they were used as a baseline for the random forest model originally. The metric that we will use to compare our models to the baseline results will be the aforementioned F1 score of each experiment.

\subsection{Results}

\begin{table}[!htb]
\captionsetup{size=footnotesize}
\caption{Experimental Results} \label{tab:results}
\setlength\tabcolsep{0pt} % let LaTeX compute intercolumn whitespace
\footnotesize\centering
\smallskip 
\begin{tabular*}{\linewidth}{@{\extracolsep{\fill}}lllll}
Model & Chunk Size & Precision & Recall & F1 \\
\hline
Kamps (Init.) & 1 Hour & 15.6\% & 96.7\% & 26.8\% \\
Kamps (Bal.) & 1 Hour & 38.4\% & 93.5\% & 54.4\% \\
Kamps (Strict) & 1 Hour & 50.1\% & 75.0\% & 60.5\% \\
RF & 5 Secs & 97.7\% & 71.6\% & 82.6 $\pm 0.0\%$\\
RF & 15 Secs & 98.0\% & 81.9\% & 89.2 $\pm 0.0\%$\\
RF & 25 Secs & 94.5\% & 83.8\% & 88.8 $\pm 0.0\%$\\
C-LSTM & 5 Secs & 91.2\% & 77.5\% & 83.7 $\pm 1.0\%$\\
C-LSTM & 15 Secs & 94.2\% & 84.9\% & 89.3 $\pm 0.4\%$\\
C-LSTM & 25 Secs & 94.2\% & 85.0\% & \textbf{89.3} $\pm 0.5\%$ \\
Anom. Trans. & 5 Secs & 91.0\% & 87.7\% & \textbf{89.3} $\pm 0.4\%$\\
Anom. Trans. & 15 Secs & 93.0\% & 94.2\% & \textbf{93.6} $\pm 0.8\%$\\
Anom. Trans. & 25 Secs & 88.4\% & 90.0\% & 89.2 $\pm 0.3\%$\\
\end{tabular*}
\caption*{\\
To show the variance of our models, the F1 metrics are followed by a 95\% confidence interval computed over $n=10$ runs. Variance is measured after our efforts to maintain reproducability via setting seeds. The random forest (RF) model's lack of variance is due to our setting seeds in a similar manner to \cite{laMorgia9209660}. Since the Kamps model is statistical in nature, it does not have variance across runs.}
\end{table}
\todo{Gireesh} %describe our results here as shown in the table preceding, and show some basis analysis/explanation as to why the results are expected
Both deep learning models were able to beat all previous classical and statistical approaches by a statistically significant margin ($p < 0.025$) except for the C-LSTM model on the 15-second chunked dataset, which matched previous results. This demonstrates the effectiveness of deep learning in this previously unexplored area. State-of-the-art results across all models are bolded in our table. On average, we found that predictions using the 5-second chunked dataset are much less accurate than those on the 15-second and 25-second chunked dataset, which suggests that predicting anomalies using smaller chunk sizes corresponds to a harder problem in general. This corroborates findings from previous works \cite{laMorgia9209660}.

A comparison of the two deep learning models that we have employed shows that the transformer was able to beat the LSTM in all but the 25-second chunk size. This is unsurprising for two reasons: firstly, transformers are state-of-the-art for time-series anomaly detection problems to begin with, and secondly, a basic transformer design without the use of the Anomaly Transformer novelties already outperforms the random forest model results -- albeit just barely.

We believe that the theoretical backing for the relative performance of the LSTM and Anomaly Transformer lies in the fact that for 5-second and 15-second chunk sizes, longer sequences are required to capture the same amount of information on average when compared to the larger 25-second chunks. Consequently, this implies a drop in the amount of temporal information present in the input for smaller chunk sizes, given a fixed segment length $s=15$ as was used in this paper. We hypothesize that this difference is responsible for the LSTM's faster drop in efficacy at smaller chunk sizes. However, more experiments must be done to reveal our models' sensitivity to the aggregation chunk size of the dataset before confirming this.

\section{Conclusion and Future Work}
\todo{viswanath} This paper studies the application of deep learning models to the crypto fraud detection problem space. We propose two separate models that can each reach state-of-the-art performance on the data available: the C-LSTM, and the Anomaly Transformer. Our results consequently show that both LSTMs and Transformers have the ability to outperform both classical ML models and statistical models on this dataset with relatively little effort. Future work includes fine-tuning these models to better-account for the volatility generally found in crypto, and exploring the potential for these models to detect pumps ahead of time as opposed to when they begin. Furthermore, the dataset provided to the community by \cite{laMorgia9209660}, while incredibly useful, has been preprocessed to be optimized in some ways for use by their random forest model; for instance, an LSTM could make use of the actual price of an asset (as opposed to the average percent change in price) in ways that a random forest model cannot. Therefore, the release of the raw dataset, if possible, would push the capabilities of deep learning models in this domain even further.

% \subsection{References}

\bibliography{anthology,acl2020}

\begin{thebibliography}{19}
\expandafter\ifx\csname natexlab\endcsname\relax\def\natexlab#1{#1}\fi

\bibitem[{Brown et~al.(2018)Brown, Tuor, Hutchinson, and
  Nichols}]{transformerbrown2018recurrent}
Andy Brown, Aaron Tuor, Brian Hutchinson, and Nicole Nichols. 2018.
\newblock Recurrent neural network attention mechanisms for interpretable
  system log anomaly detection.
\newblock In \emph{Proceedings of the First Workshop on Machine Learning for
  Computing Systems}, pages 1--8.

\bibitem[{Coffeezilla(2021)}]{coffeezilla_2021}
Coffeezilla. 2021.
\newblock \href {https://www.youtube.com/watch?v=ehDvr5vVGPg} {I joined a pump
  and dump scheme so you don't have to - youtube}.

\bibitem[{Cumming et~al.(2019)Cumming, Johan, and
  Pant}]{regulation_jrfm12030126}
Douglas~J. Cumming, Sofia Johan, and Anshum Pant. 2019.
\newblock \href {https://doi.org/10.3390/jrfm12030126} {Regulation of the
  crypto-economy: Managing risks, challenges, and regulatory uncertainty}.
\newblock \emph{Journal of Risk and Financial Management}, 12(3).

\bibitem[{Hamrick et~al.(2021)Hamrick, Rouhi, Mukherjee, Feder, Gandal, Moore,
  and Vasek}]{hamrick}
J.T. Hamrick, Farhang Rouhi, Arghya Mukherjee, Amir Feder, Neil Gandal, Tyler
  Moore, and Marie Vasek. 2021.
\newblock \href {https://doi.org/https://doi.org/10.1016/j.ipm.2021.102506} {An
  examination of the cryptocurrency pump-and-dump ecosystem}.
\newblock \emph{Information Processing \& Management}, 58(4):102506.

\bibitem[{Hasanin and Khoshgoftaar(2018)}]{8424689}
Tawfiq Hasanin and Taghi Khoshgoftaar. 2018.
\newblock \href {https://doi.org/10.1109/IRI.2018.00018} {The effects of random
  undersampling with simulated class imbalance for big data}.
\newblock In \emph{2018 IEEE International Conference on Information Reuse and
  Integration (IRI)}, pages 70--79.

\bibitem[{Kamps and Kleinberg(2018)}]{kamps}
J.~Kamps and B.~Kleinberg. 2018.
\newblock \href {https://doi.org/10.1186/s40163-018-0093-5} {To the moon:
  defining and detecting cryptocurrency pump-and-dumps}.
\newblock \emph{Crime Sci}.

\bibitem[{Keogh et~al.(2021)Keogh, Roy, U, and A.}]{nasadataset}
Eamonn~J. Keogh, Taposh Roy, Naik U, and Agrawal A. 2021.
\newblock \href {https://compete.hexagon-ml.com/practice/competition/39/}
  {Multi-dataset time-series anomaly detection competition, competition of
  international conference on knowledge discovery \& data mining}.

\bibitem[{Khatri(2021)}]{cryptostats_khatri_2021}
Yogita Khatri. 2021.
\newblock \href
  {https://web.archive.org/web/20220223151030/https://www.theblockcrypto.com/linked/128526/centralized-crypto-exchanges-14-trillion-trading-volume-2021}
  {Crypto exchanges saw over \$14 trillion in trading volume this year}.

\bibitem[{Kim and Cho(2018)}]{clstmKIM201866}
Tae-Young Kim and Sung-Bae Cho. 2018.
\newblock \href {https://doi.org/https://doi.org/10.1016/j.eswa.2018.04.004}
  {Web traffic anomaly detection using c-lstm neural networks}.
\newblock \emph{Expert Systems with Applications}, 106:66--76.

\bibitem[{Kwon et~al.(2018)Kwon, Natarajan, Suh, Kim, and
  Kim}]{kwon2018empiricalconv}
Donghwoon Kwon, Kathiravan Natarajan, Sang~C Suh, Hyunjoo Kim, and Jinoh Kim.
  2018.
\newblock An empirical study on network anomaly detection using convolutional
  neural networks.
\newblock In \emph{ICDCS}, pages 1595--1598.

\bibitem[{La~Morgia et~al.(2020)La~Morgia, Mei, Sassi, and
  Stefa}]{laMorgia9209660}
Massimo La~Morgia, Alessandro Mei, Francesco Sassi, and Julinda Stefa. 2020.
\newblock \href {https://doi.org/10.1109/ICCCN49398.2020.9209660} {Pump and
  dumps in the bitcoin era: Real time detection of cryptocurrency market
  manipulations}.
\newblock In \emph{2020 29th International Conference on Computer
  Communications and Networks (ICCCN)}, pages 1--9.

\bibitem[{Lai et~al.(2022)Lai, Zha, Xu, Zhao, Wang, and
  Hu}]{neuripstimeseriesbenchmark}
Kwei-Herng Lai, Daochen Zha, Junjie Xu, Yue Zhao, Guanchu Wang, and Xia Hu.
  2022.
\newblock Revisiting time series outlier detection: Definitions and benchmarks.

\bibitem[{Leangarun et~al.(2018)Leangarun, Tangamchit, and
  Thajchayapong}]{stock_pandd_8628777}
Teema Leangarun, Poj Tangamchit, and Suttipong Thajchayapong. 2018.
\newblock \href {https://doi.org/10.1109/SSCI.2018.8628777} {Stock price
  manipulation detection using generative adversarial networks}.
\newblock In \emph{2018 IEEE Symposium Series on Computational Intelligence
  (SSCI)}, pages 2104--2111.

\bibitem[{Malhotra et~al.(2016)Malhotra, Ramakrishnan, Anand, Vig, Agarwal, and
  Shroff}]{malhotra2016lstm}
Pankaj Malhotra, Anusha Ramakrishnan, Gaurangi Anand, Lovekesh Vig, Puneet
  Agarwal, and Gautam Shroff. 2016.
\newblock Lstm-based encoder-decoder for multi-sensor anomaly detection.
\newblock \emph{arXiv preprint arXiv:1607.00148}.

\bibitem[{Su et~al.(2019)Su, Zhao, Niu, Liu, Sun, and
  Pei}]{10.1145/3292500.3330672}
Ya~Su, Youjian Zhao, Chenhao Niu, Rong Liu, Wei Sun, and Dan Pei. 2019.
\newblock \href {https://doi.org/10.1145/3292500.3330672} {Robust anomaly
  detection for multivariate time series through stochastic recurrent neural
  network}.
\newblock In \emph{Proceedings of the 25th ACM SIGKDD International Conference
  on Knowledge Discovery \& Data Mining}, KDD '19, page 2828–2837, New York,
  NY, USA. Association for Computing Machinery.

\bibitem[{Twomey and Mann(2020)}]{twomey2020fraud}
David Twomey and Andrew Mann. 2020.
\newblock Fraud and manipulation within cryptocurrency markets.
\newblock \emph{Corruption and fraud in financial markets: malpractice,
  misconduct and manipulation}, 624.

\bibitem[{Vaswani et~al.(2017)Vaswani, Shazeer, Parmar, Uszkoreit, Jones,
  Gomez, Kaiser, and Polosukhin}]{attention_viswani}
Ashish Vaswani, Noam Shazeer, Niki Parmar, Jakob Uszkoreit, Llion Jones,
  Aidan~N. Gomez, Lukasz Kaiser, and Illia Polosukhin. 2017.
\newblock \href {http://arxiv.org/abs/1706.03762} {Attention is all you need}.
\newblock \emph{CoRR}, abs/1706.03762.

\bibitem[{Xu et~al.(2021)Xu, Wu, Wang, and Long}]{anomalytransformer}
Jiehui Xu, Haixu Wu, Jianmin Wang, and Mingsheng Long. 2021.
\newblock \href {http://arxiv.org/abs/2110.02642} {Anomaly transformer: Time
  series anomaly detection with association discrepancy}.
\newblock \emph{CoRR}, abs/2110.02642.

\bibitem[{Yang et~al.(2020)Yang, Wang, and Wang}]{yang9382507}
Wenjie Yang, Ruofan Wang, and Bofan Wang. 2020.
\newblock \href {https://doi.org/10.1109/MSIEID52046.2020.00029} {Detection of
  anomaly stock price based on time series deep learning models}.
\newblock In \emph{2020 Management Science Informatization and Economic
  Innovation Development Conference (MSIEID)}, pages 110--114.

\end{thebibliography}
\bibliographystyle{acl_natbib}

\appendix

%\section{Appendices}
\iffalse
\begin{table*}[!htb]
\captionsetup{size=footnotesize}
\caption{Results} \label{tab:freq}
\setlength\tabcolsep{10pt} % let LaTeX compute intercolumn whitespace
\footnotesize\centering
something here

\smallskip 
\begin{tabular*}{\textwidth}{@{\extracolsep{\fill}}c|ccc|ccc|ccc}
&
\multicolumn{3}{c|}{5S-Chunked Data} &
\multicolumn{3}{c|}{15S-Chunked Data} &
\multicolumn{3}{c}{25S-Chunked Data} \\
\hline
Model &
Precision & Recall & F1 & Precision & Recall & F1 & Precision & Recall & F1 \\
\hline
RF & 1 & 2 & 3 & 1 & 2 & 3 &  1 & 2 & 3 \\

C-LSTM & 1 & 2 & 3 & 1 & 2 & 3 & 1 & 2 & 3 \\

Anomaly Transformer & 1 & 2 & 3 & 1 & 2 & 3 & 1 & 2 & 3 \\
\end{tabular*}
\end{table*}
\fi
\end{document}